\documentclass[conference]{IEEEtran}
\IEEEoverridecommandlockouts
\usepackage{cite}
\usepackage{amsmath,amssymb,amsfonts}
\usepackage{algorithmic}
\usepackage{algorithm}
\usepackage{graphicx}
\usepackage{textcomp}
\usepackage{xcolor}
\def\BibTeX{{\rm B\kern-.05em{\sc i\kern-.025em b}\kern-.08em
    T\kern-.1667em\lower.7ex\hbox{E}\kern-.125emX}}

\usepackage{rotating}
\usepackage{float}
 \usepackage[nameinlink]{cleveref}

\begin{document}

\title{Iterative Nadaraya-Watson  Distribution Transfer for Colour Grading\\ \thanks{This work is partly funded by a scholarship from Umm Al-Qura University, Saudi Arabia, and in part by a research grant from Science Foundation Ireland (SFI) under the Grant Number 15/RP/2776, and the ADAPT Centre for Digital Content Technology (www.adaptcentre.ie) that is funded under the SFI Research Centres Programme (Grant 13/RC/2106) and is co-funded under the European Regional Development Fund.}
}

\author{\IEEEauthorblockN{Hana Alghamdi}
\IEEEauthorblockA{\textit{School of Computer Science \& Statistics} \\
\textit{Trinity College Dublin}\\
Dublin, Ireland \\
alghamdh@tcd.ie}
\and
\IEEEauthorblockN{Rozenn Dahyot}
\IEEEauthorblockA{\textit{School of Computer Science \& Statistics} \\
\textit{Trinity College Dublin}\\
Dublin, Ireland  \\
rozenn.dahyot@tcd.ie}

}

\maketitle

\begin{abstract}

We propose a new method with Nadaraya-Watson that maps one N-dimensional distribution to another taking into account available information about correspondences. We extend the 2D/3D problem to higher dimensions by encoding overlapping neighborhoods of data points and solve the high dimensional problem in 1D space using an iterative projection approach.
To show potentials of this mapping, we apply it  to colour transfer between two images that exhibit overlapped scene. Experiments show quantitative and qualitative improvements over previous state of the art colour transfer methods.
\end{abstract}


\section{Introduction}

\begin{figure}[!h]
\begin{center}
 \includegraphics[width=.92\linewidth]{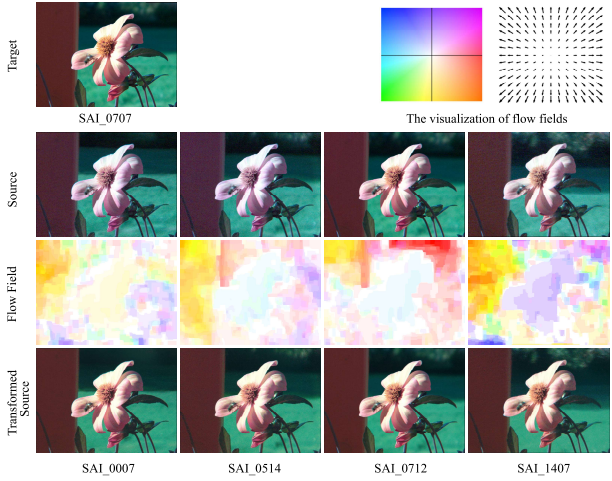}
\end{center}
   \caption{Results on  SAI light fields images. Our method \texttt{NW\_cp}  successfully corrects colour inconsistencies in the selected outer views images using the centre view image (SAI\_0707) as  the target image. The flow fields  show the motion changes between the target and each source image. The top row on the right shows flow field visualization based on the code in \cite{Baker2007}: each pixel denotes a flow vector where the orientation and magnitude are represented by the hue and saturation of the pixel, respectively.}
\label{fig:LF}
\end{figure}

Colour variations between photographs often happen due to illumination changes, using different cameras, different in-camera settings or due to tonal adjustments of the users. Colour transfer methods have been developed to transform a source colour image into a specified target colour image to match colour statistics or eliminate colour variations between different photographs. Applications of colour transfer in image processing problems are various,  ranging from  colour correction for image mosaicing and stitching  \cite{brown2007automatic}, to colour enhancement and style manipulation for artistic design applications \cite{hwang2014color}.
An example of application of colour transfer is illustrated in  Figure \ref{fig:LF}: light fields  have become a major research topic and among the different methods used to capture a light field are the lenslet cameras that  extract sub-aperture images (SAI), each with a very wide depth of field and representing different viewpoints of the scene. However, as can be seen, the extracted views suffer from a number of artifacts such as colour discrepancies \cite{GroganCVMP2019,MatysiakTIP2020}. 

\begin{figure}[!h]
\includegraphics[width=\linewidth]{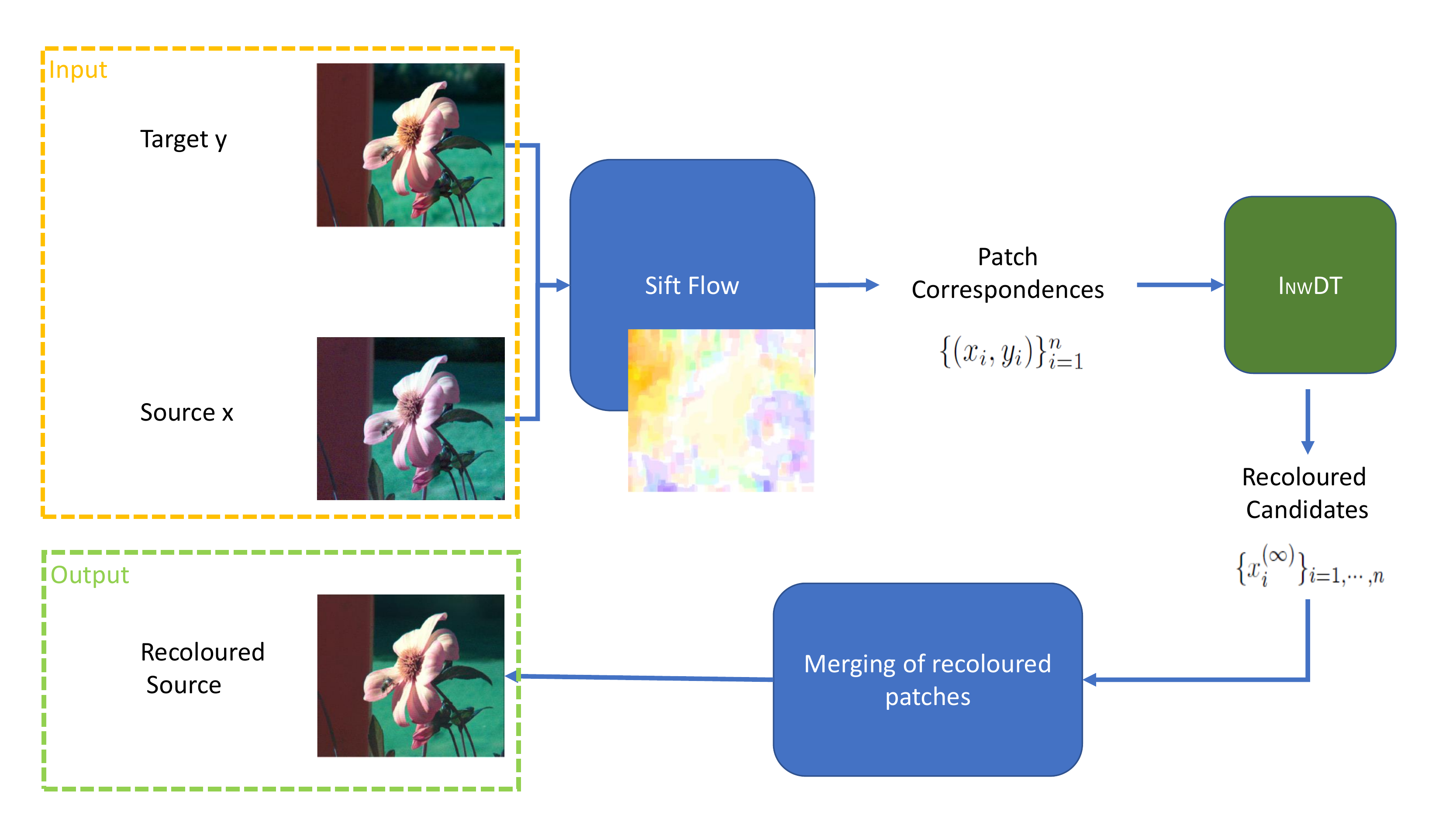}
\caption{Proposed Pipeline: Following  \cite{alghamdi2020patch}, correspondences between target and source are found using SIFT flow \cite{SIFTflow2010}. These correspondences are used in our proposed INWDT algorithm (cf. Alg. \ref{Alg:IDT:NW}) to compute recoloured candidates that are then merged using the same process as \cite{Alghamdi2019} to compose the recoloured source image.}
\label{fig:pipeline}
\end{figure}

When target and source images are from the same scene, correspondences can be found to guide the process for recolouring \cite{GroganCVIU19}. The SIFT flow algorithm \cite{SIFTflow2010} is well suited  for matching densely sampled, pixel-wise SIFT features between the two images \cite{alghamdi2020patch} and is used in our proposed pipeline (cf. Fig. \ref{fig:pipeline}). In this paper, we propose to use these correspondences between source and target images for performing colour transfer with a new algorithm (cf. Sec. \ref{sec:INWDT}) noted INWDT (cf. Fig. \ref{fig:pipeline}).
We compare our approach  against state of the art techniques for colour transfer  \cite{Pitie_CVIU2007,hwang2014color,bellavia2018dissecting,GroganCVIU19,Alghamdi2019}
(Sec. \ref{sec:experiments}) and show competitive results, both quantitatively and qualitatively. 

\section{Iterative Nadaraya-Watson Distribution Transfer}
\label{sec:INWDT}

We explain our INWDT algorithm  (Alg. \ref{Alg:IDT:NW})
in Section \ref{sec:NW}. It  is derived from  the Iterative Distribution Transfer (IDT) algorithm originally proposed by Pitie et al. \cite{Pitie_CVIU2007} as a solution to optimal transport in N-dimensional spaces. 
Our algorithm is  part of our overall pipeline (Fig. \ref{fig:pipeline})  that is explained in Section \ref{sec:overall:pipeline}.  

\begin{algorithm}[!h]
\begin{algorithmic}[1]
	\STATE{\textbf{Input:}  Target  \& Source datasets with correspondences  $\{(x_i,y_i)\}_{i=1}^n$, samples of r.v. $(x,y) \in \mathbb{R}^d\times \mathbb{R}^d$ }
	\STATE{\textbf{Initialisation:}  $k\leftarrow 0$ and   $\forall i,\ x_i^{(0)}\leftarrow x_i$}
	\REPEAT 
	 \STATE{Generate $N$ random unit vectors in $\mathbb{R}^d$ stored in matrix $\mathrm{R} = [e_1, ... , e_N]$}
	 \FOR{$j=1$ to $N$}
	 \STATE{Compute projections $\forall i$ $u_i=e_j^T x_i^{(k)}$ and $v_i=e_j^T y_i$}
	 \STATE{Compute 1D NW estimate $v=\phi_j(u)$ with  $\lbrace (u_i,v_i)\rbrace$}
	 \ENDFOR
	 \STATE{Remap the source dataset
	 $$x^{(k+1)}_i=x^{(k)}_i+\mathrm{R}^{-1} \left(\begin{array}{c} 
	 \phi_1(e_1^T x_i^{(k)})-e_1^T x_i^{(k)}\\
	 \vdots\\
	 \phi_N(e_N^T x_i^{(k)})-e_N^T x_i^{(k)}\\
	 \end{array}\right)$$
	 }
	 \STATE $k\leftarrow k+1$
	 \UNTIL{convergence $\mathcal{L}_2(p_{x},p_{y})\rightarrow 0$ (noted $k\equiv \infty$)}
	\STATE{\textbf{Result:} With the recoloured patches $\lbrace x_i^{(\infty)}\rbrace_{i=1,\cdots,n}$, the final one-to-one mapping $\Phi$ in $\mathbb{R}^d$ is given by $x_i\rightarrow \Phi(x_i)=x_i^{(\infty)},\ \forall i$.}
	\end{algorithmic}
	\caption{Iterative Nadaraya-Watson  Distribution Transfer}
	\label{Alg:IDT:NW}
	\end{algorithm}

\subsection{Nadaraya-Watson Vs Optimal transport solution in 1D}
\label{sec:NW}

\begin{figure}[!h]
\begin{center}
\includegraphics[width = .8\linewidth]{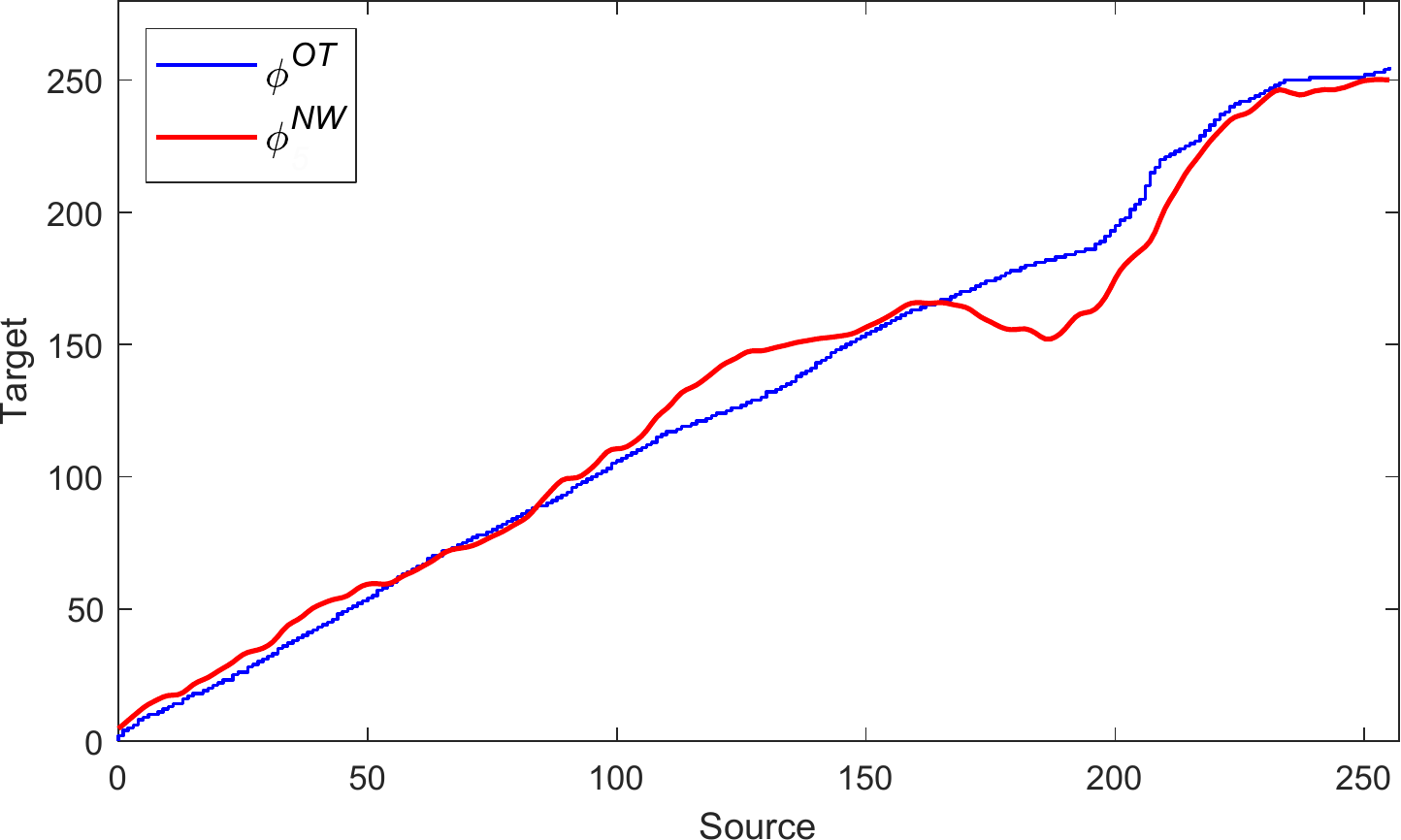} 
\end{center}
   \caption{Illustration of the non-parametric Nadaraya-Watson mapping function ($\phi^{NW}$) that accounts for correspondences compared with the strictly increasing Optimal transport function ($\phi^{OT}$) that does not take into account the correspondences.}
\label{fig:OTNW_h}
\end{figure}

The IDT algorithm \cite{Pitie_CVIU2007}  proposes to project two multidimensional independent datasets $\lbrace x_i \rbrace $ and $\lbrace y_i \rbrace$ sampled for two random vectors $x\in \mathbb{R}^d$ and $y\in \mathbb{R}^d$ with respective distributions $p_x$ and $p_y$,    in a one dimensional space (cf. line 6 in Alg. \ref{Alg:IDT:NW}). This projection creates two datasets  $\lbrace u_i \rbrace $ and $\lbrace v_i \rbrace$ whose cumulative distributions $P_u$ and $P_v$ are registered using the 1D optimal transport solution in IDT \cite{Pitie_CVIU2007}:
\begin{equation}
\phi^{OT}(u)=P_v^{-1} \circ P_u(u)  
\label{eq:OT}
\end{equation}
We replace $\phi^{OT}$ used in IDT by the Nadaraya-Watson (NW) estimate (cf. Alg.  \ref{Alg:IDT:NW} line 7) taking advantage of the correspondences $\lbrace (x_i,y_i) \rbrace$ giving correspondences $\lbrace (u_i,v_i) \rbrace$ in the projective space:
\begin{equation}
\phi^{NW}(u)= \frac{\sum_{i=1}^n v_i \ K_h(u-u_i)}{\sum_{i=1}^n  K_h(u-u_i) } \simeq \mathbb{E}_{p_{u|v}}[u|v]
\label{eq:NW}
\end{equation}
The NW estimator computes the estimate of an expectation of $u$ given $v$  using a kernel (e.g. Gaussian) with bandwidth $h$. This bandwidth controls  the smoothness of the estimated function $\phi^{NW}$. 
Fig \ref{fig:OTNW_h} presents the two estimates $\phi^{OT}$ and $\phi^{NW}$
estimated as part of one iteration of our algorithm. Not having correspondences,  
$\phi^{OT}$ is by definition (Eq. \ref{eq:OT}) a strictly increasing function, whereas  $\phi^{NW}$ provides a smooth non-monotonic mapping function $u$ to $v$.  

\subsection{Pipeline}
\label{sec:overall:pipeline}

\subsubsection{Patch correspondences}
 We use the same process explained in  \cite{alghamdi2020patch} to create a set of corresponding patches $\lbrace (x_i,y_i)\rbrace$ (patch size neighborhood $m\times m$, creating data dimension $d=m\times m \times 3$), where each pair corresponds to  two  vectors $x_i$ and $y_i$.   SIFT flow motion estimation \cite{SIFTflow2010} is used to define the  correspondences $\{(x_i,y_i)\}_{i=1}^n$.  We consider likewise patches containing only colour information of a pixel neighborhood, and as an alternative definition, patches define with pixel location information in addition to colour information \cite{alghamdi2020patch}.  
 



\subsubsection{Iterative Nadaraya-Watson Distribution Transfer}
\begin{figure}[!h]
\begin{center}
\includegraphics[width = 1\linewidth, height= 1.2\linewidth]{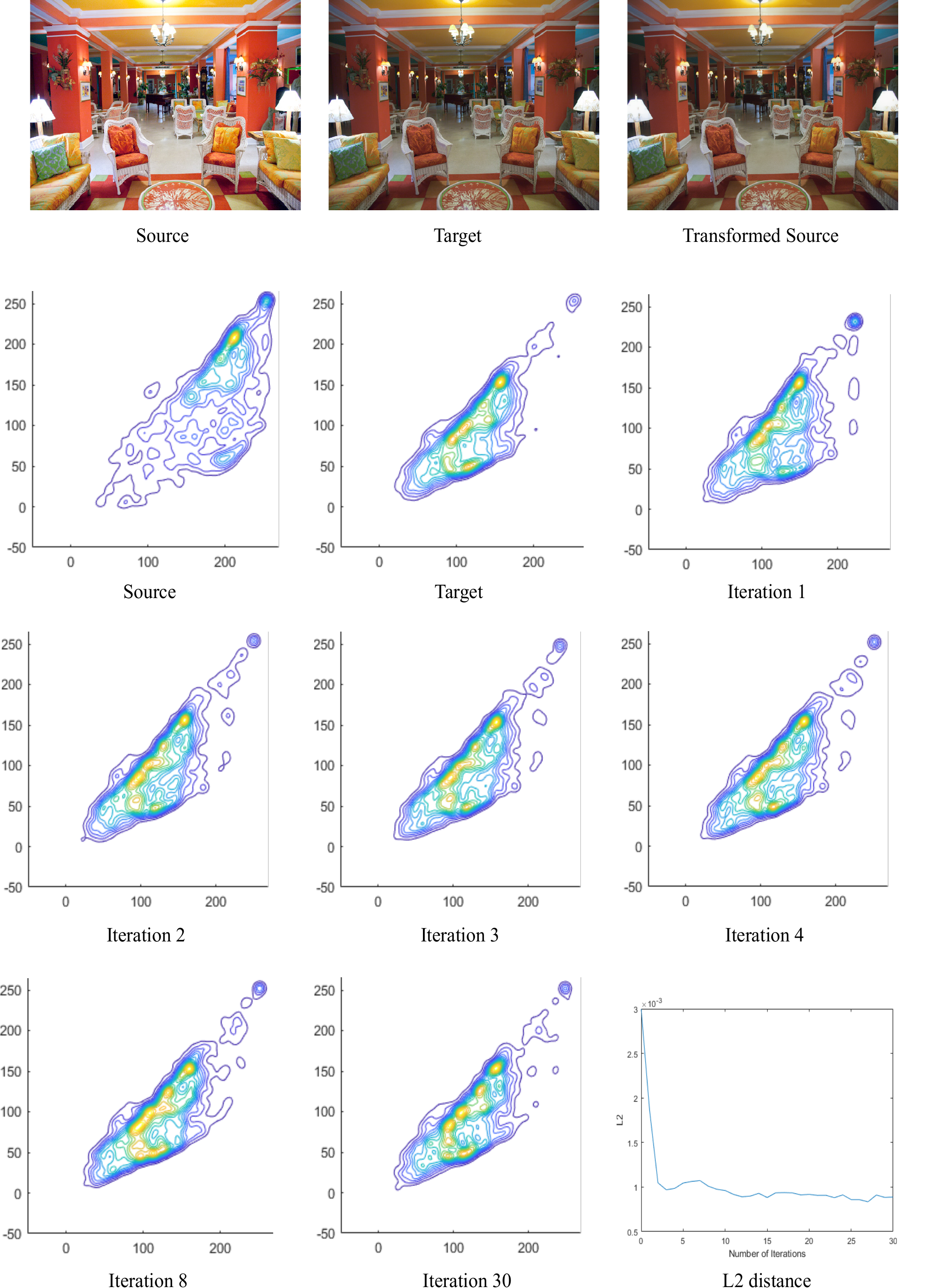} 
\end{center}
   \caption{Example of pdf of the transferred source patches projected in 2D space (RG). The patch size chosen is $1\times 1$ and  only the colour information is used $d=3$ (space RGB). The $\mathcal{L}_2$ distance \cite{GroganCVIU19} is computed at each iteration to illustrate the convergence of the original distribution to the target one by our transfer method.}
\label{fig:pdf}
\end{figure}

Our algorithm  outlined in Algorithm \ref{Alg:IDT:NW} is applied to our set $\{(x_i,y_i)\}_{i=1}^n$ (input) to compute recoloured patches $\{x_i^{(\infty)}\}_{i=1}^n$.
Fig. \ref{fig:pdf} illustrates several iterations $k$ of our algorithm visualised in 2D space. 

\subsubsection{Merge recoloured candidates.}
Because the same pixel is present in overlapping patches $\{x_i^{(\infty)}\}_{i=1}^n$,  the average colour value from all the candidates is selected for recolouring  \cite{alghamdi2020patch,Alghamdi2019}.


\section{Experimental Assessment}
 \label{sec:experiments}
 
We provide quantitative and qualitative evaluations to validate both of our NW solutions - using colour patches only, annotated in the results as \texttt{NW\_c}, and using colour patches with  pixel  location  information, annotated as \texttt{NW\_cp}.
We compare our methods to different state of the art colour transfer methods  noted  \texttt{IDT} \cite{Pitie_CVIU2007}, \texttt{PMLS} \cite{hwang2014color}, \texttt{GPS/LCP} and \texttt{FGPS/LCP}  \cite{bellavia2018dissecting}, \texttt{L2} \cite{GroganCVIU19}, \texttt{PCT\_OT} \cite{Alghamdi2019} and \texttt{OT\_NW} \cite{alghamdi2020patch}. In these evaluations we use image pairs with
similar content from an existing dataset provided by Hwang et al. \cite{hwang2014color}. The dataset includes registered pairs of images (source and target) taken with different cameras, different in-camera settings, and different illuminations and recolouring styles.

\subsection{Colour space and parameters settings}
We use  the RGB colour space and  we found a patch size of $3 \times 3$ captures enough of a pixel's neighbourhood.  For our \texttt{NW\_cp} version, each pixel is represented by its 3D RGB colour values  and its 2D pixel position (i.e 5D). The patches with combined colour and spatial features create a vector in 45 dimensions ($d=3 \times 3\times 5=45$). For \texttt{NW\_c}, pixel position is not accounted for, and only RGB colours are used ($d=3 \times 3\times 3=27$).  We experimented with different bandwidth values  (cf. Eq. \ref{eq:NW}) and we found a fix value of $h=5$ gives best results.

\subsection{Evaluation metrics}
To quantitatively assess the recolouring results,  four metrics are used: peak signal to noise ratio (PSNR) \cite{salomon2004data}, structural similarity index (SSIM) \cite{wang2004image}, colour image difference (CID) \cite{preiss2014color} and feature similarity index (FSIMc) \cite{zhang2011fsim}. These metrics are often used when considering source and target images of the same content \cite{GroganCVIU19,lissner2013image, oliveira2015probabilistic, hwang2014color, bellavia2018dissecting}. Note that the results using \texttt{PMLS} were provided by the authors \cite{hwang2014color}. It has already been shown in \cite{GroganCVIU19}  that \texttt{PMLS} performs better than  two other more recent techniques using correspondences \cite{park2016, Xia2017},  
so \texttt{PMLS} is the one reported here  with \cite{alghamdi2020patch, Alghamdi2019} and \cite{GroganCVIU19} as algorithms that incorporate correspondences in their methodologies.


\subsection{Experimental Results}

\begin{figure}[h]
\centering
\includegraphics[width = .8\linewidth, height= 1.2\linewidth]{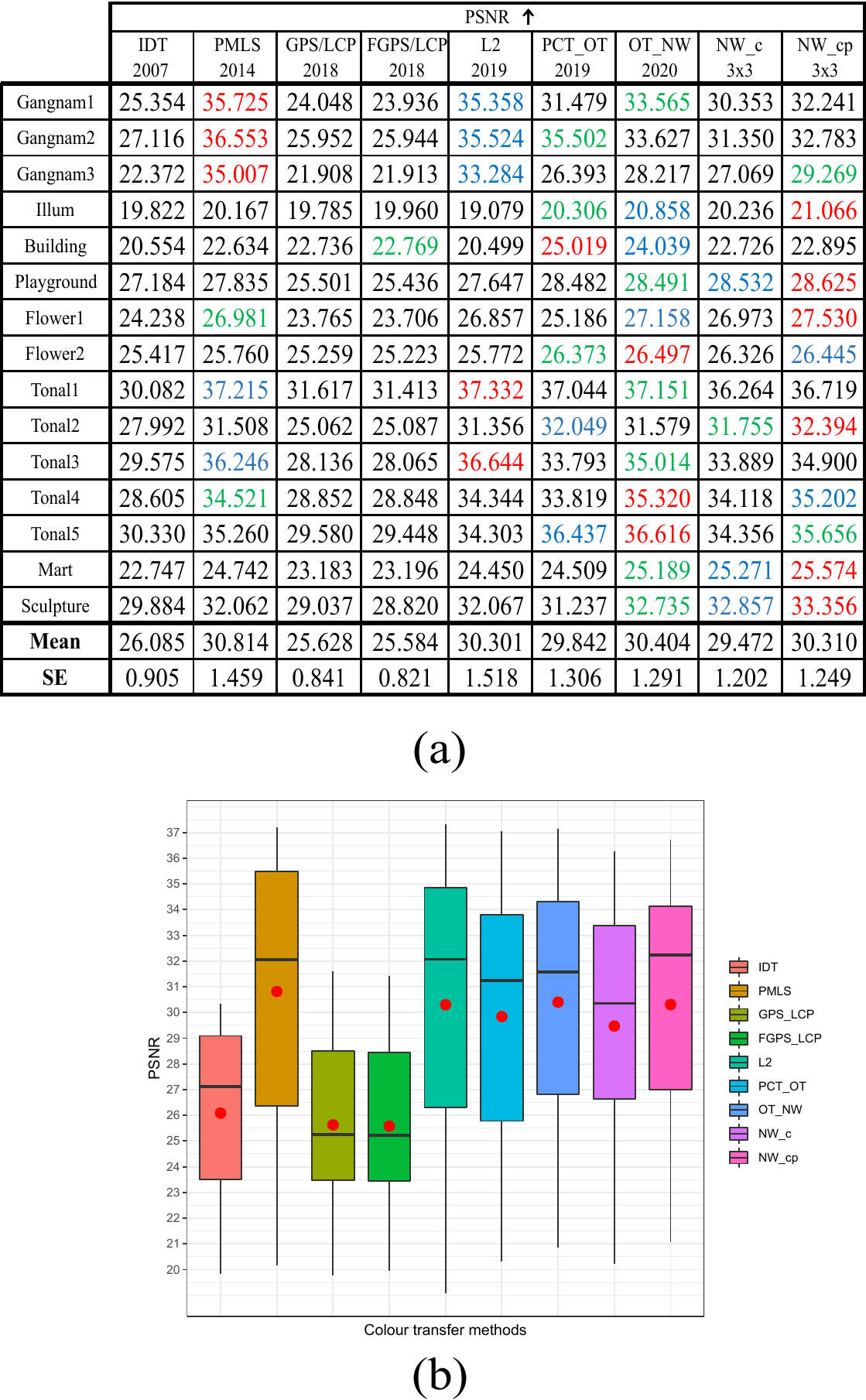}
\caption{Metric comparison, using PSNR \cite{salomon2004data}. (a) Red, blue, and green indicate $1^{st}$, $2^{nd}$, and $3^{rd}$ best performance respectively in the table (higher values are better), (b) visualized in box plot (best viewed in colour and zoomed in).}
\label{fig:psnr}
\end{figure}

\begin{figure}[h]
\centering
\includegraphics[width = .8\linewidth, height= 1.2\linewidth]{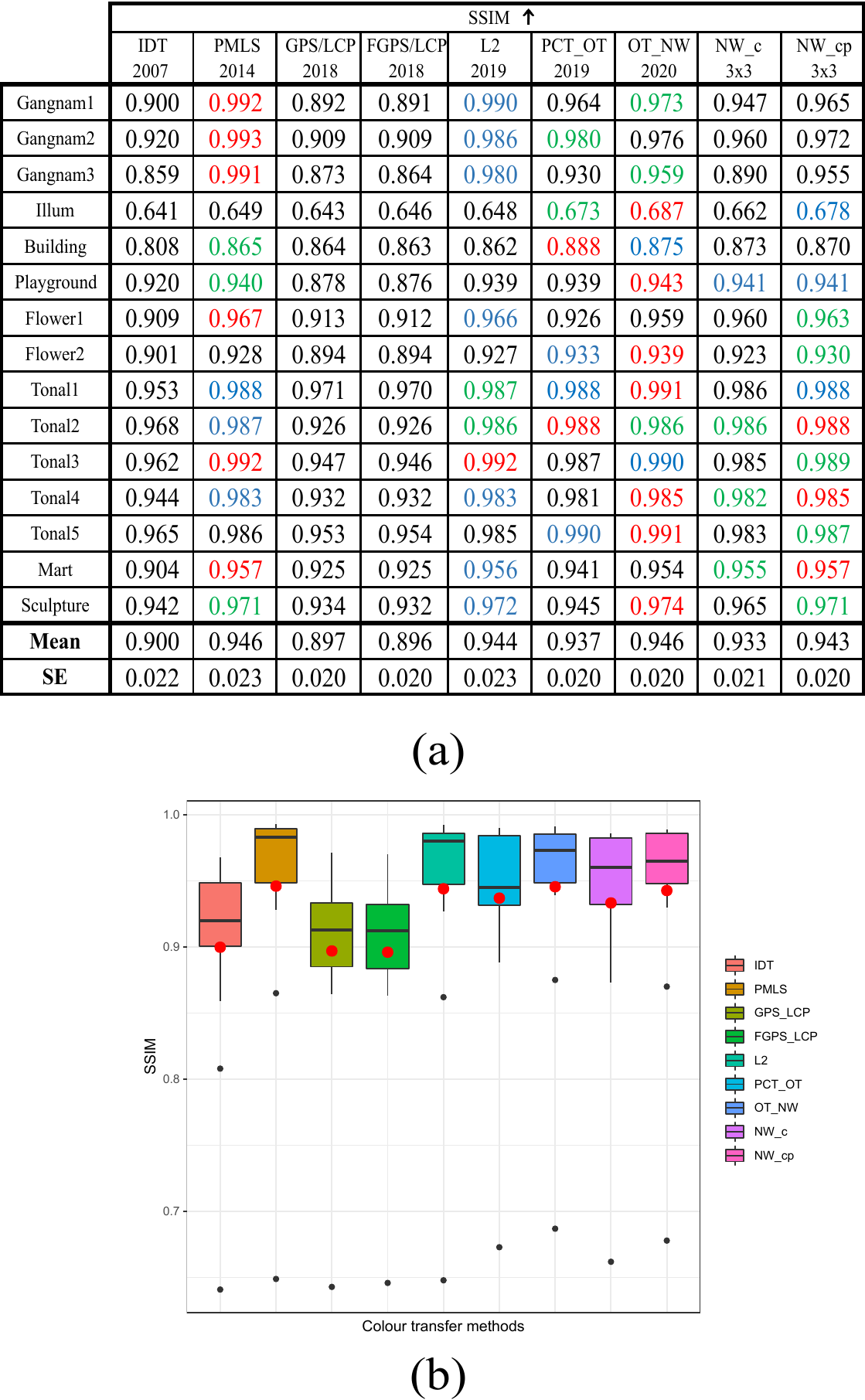}

\caption{Metric comparison, using SSIM \cite{wang2004image}. (a) Red, blue, and green indicate $1^{st}$, $2^{nd}$, and $3^{rd}$ best performance respectively in the table (higher values are better), (b) visualized in box plot (best viewed in colour and zoomed in).}
\label{fig:ssim}
\end{figure}

\begin{figure}[!ht]
\centering
\includegraphics[width = .8\linewidth, height= 1.2\linewidth]{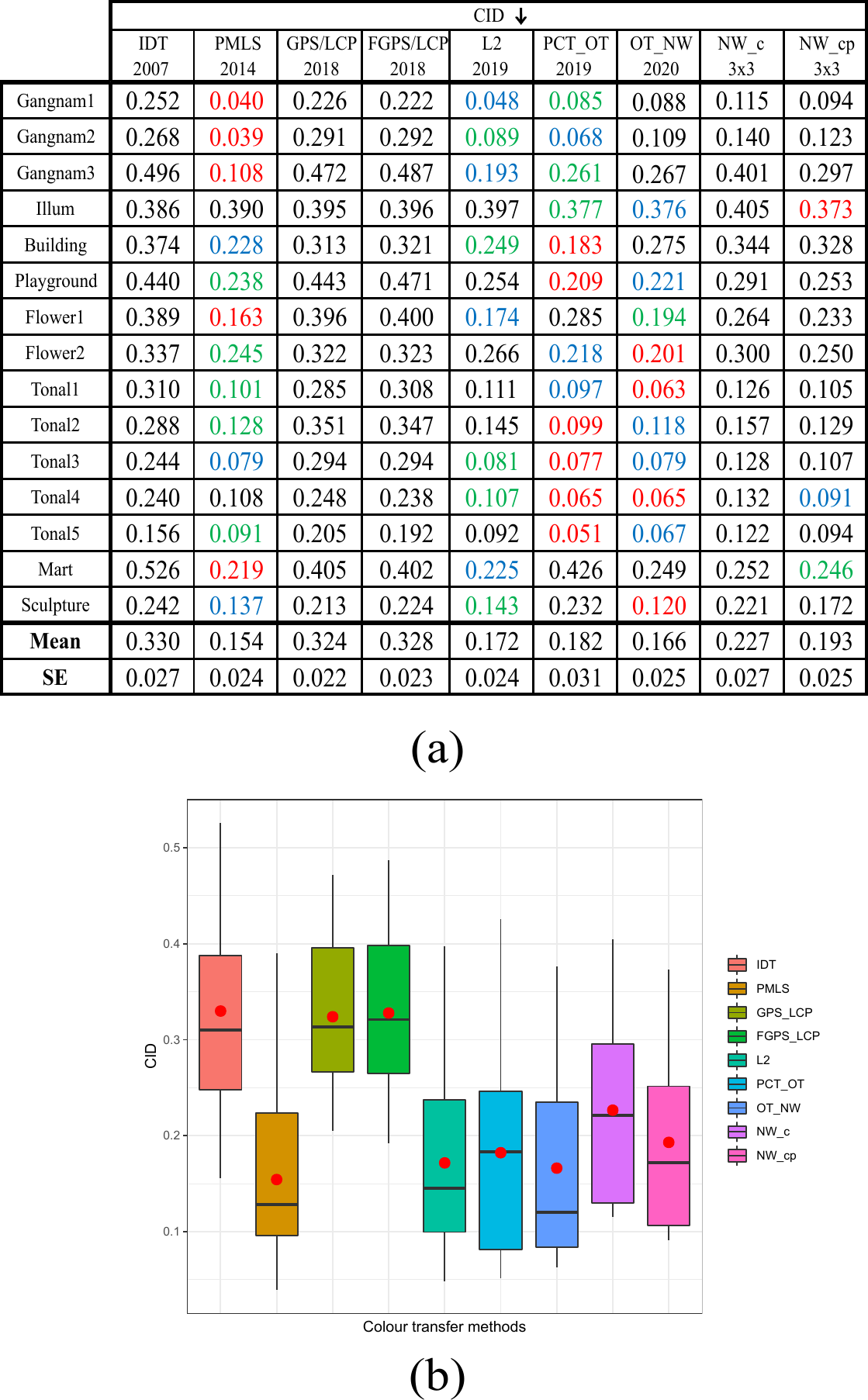}
\caption{Metric comparison, using CID \cite{preiss2014color}. (a) Red, blue, and green indicate $1^{st}$, $2^{nd}$, and $3^{rd}$ best performance respectively in the table (lower values are better), (b) visualized in box plot (best viewed in colour and zoomed in).}
\label{fig:cid}
\end{figure}

\begin{figure}[!ht]
\centering
\includegraphics[width = .8\linewidth, height= 1.2\linewidth]{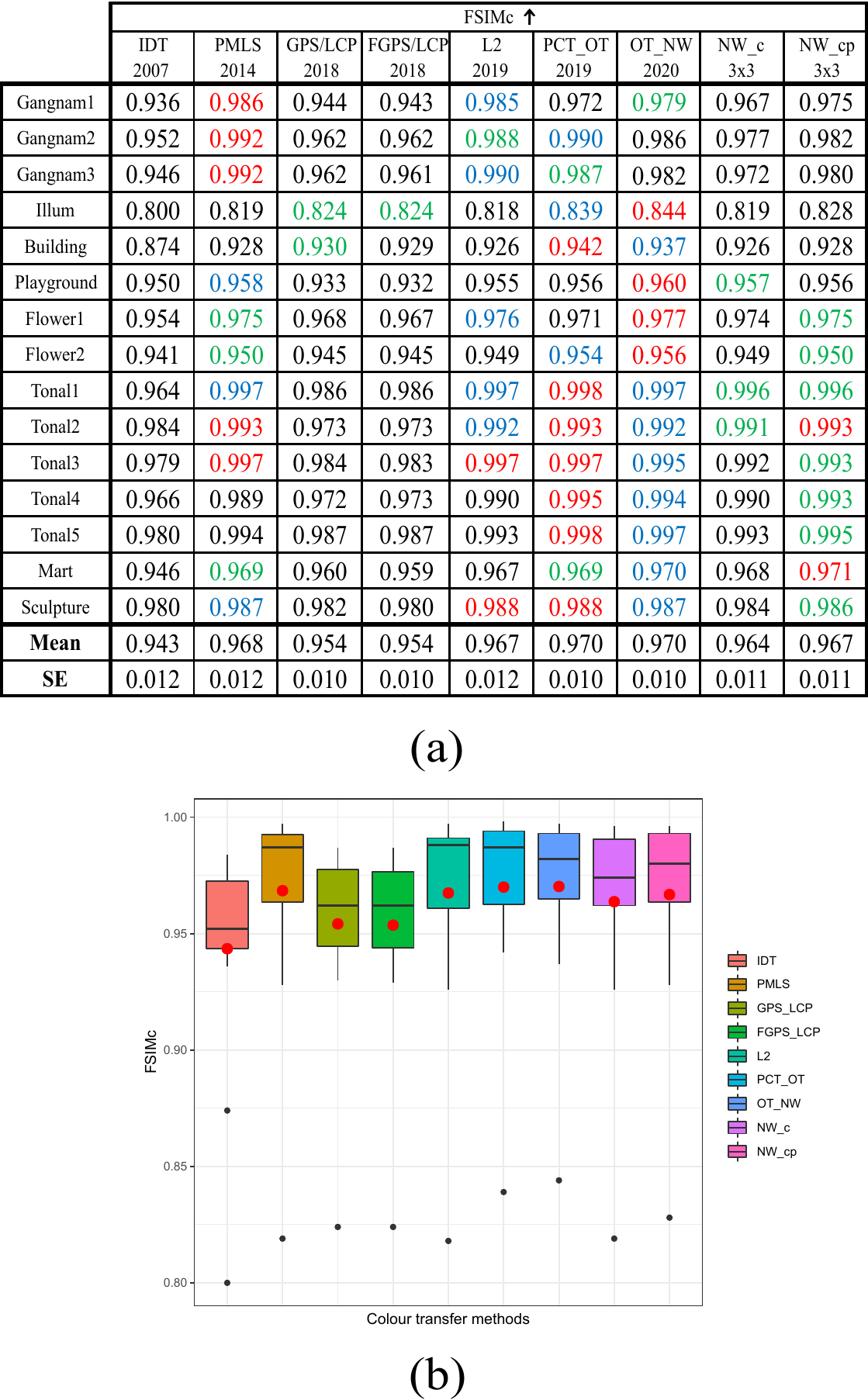}
\caption{Metric comparison, using FSIMc \cite{zhang2011fsim}. (a) Red, blue, and green indicate $1^{st}$, $2^{nd}$, and $3^{rd}$ best result respectively in the table (higher values are better), (b) visualized in box plot (best viewed in colour and zoomed in).}
\label{fig:fsimc}
\end{figure}

\Cref{fig:psnr,fig:ssim,fig:cid,fig:fsimc} show detailed tables of quantitative results with means and standard errors (SE) for each metric  alongside with box plots carrying  a lot of statistical details. The purpose of the  box plots is to visualize differences among methods and to show how close our method is to the state of the art algorithms. \Cref{fig:psnr} (b) and  \Cref{fig:fsimc} (b) shows PSNR and FSIMc metrics results respectively, although our method \texttt{NW\_cp} outperforms other state of the art methods in many cases as measured by PSNR as shown in the table, by examining the box plots in both figures we see that the methods \texttt{PMLS}, \texttt{L2}, \texttt{PCT\_OT}, \texttt{OT\_NW}, \texttt{NW\_c} and \texttt{NW\_cp} are greatly overlap with each other, the median and mean values (the mean shown as red dots in the plots) are the highest among all algorithms and  are very close in value and the whiskers length almost similar indicating similar data variation and consistency.  Similarly, \Cref{fig:ssim} (b) shows SSIM box plot, we can see that  \texttt{NW\_cp} gives a closer performance to  \texttt{OT\_NW}, \texttt{PMLS} and \texttt{L2}  scoring highest values while \texttt{PCT\_OT} greatly overlap with \texttt{NW\_c}. In \Cref{fig:cid},  CID metric shows that \texttt{NW\_cp} that combines the spatial and colour information  better than \texttt{NW\_c} in giving a similar  performance  with top methods. In conclusion, the quantitative metrics and the standard errors  show that statistically our algorithms with Nadaraya-Watson give a comparative performance  with top methods \texttt{PMLS}\cite{hwang2014color}, \texttt{L2}\cite{GroganCVIU19}, \texttt{OT\_NW} \cite{alghamdi2020patch} and \texttt{PCT\_OT} \cite{Alghamdi2019} and outperforms the rest of  the state of the art algorithms \cite{Pitie_CVIU2007,bellavia2018dissecting}.

\begin{sidewaysfigure*}[h]
    \centering
    \includegraphics[width=\textwidth] {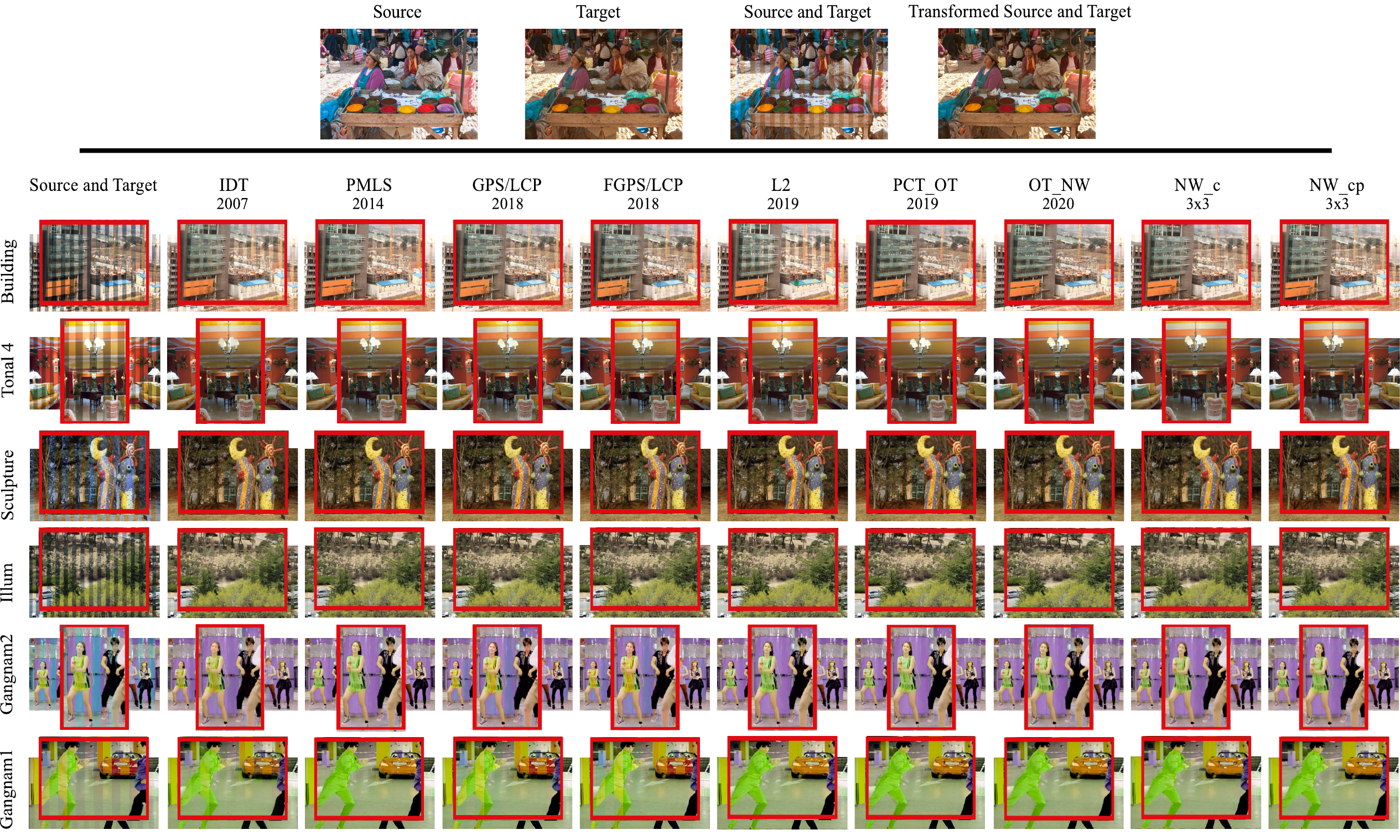}
    \caption{A close up look at some of the results generated using the \texttt{IDT} \cite{Pitie_CVIU2007}, \texttt{PMLS} \cite{hwang2014color}, \texttt{GPS/LCP} and \texttt{FGPS/LCP}  \cite{bellavia2018dissecting}, \texttt{L2} \cite{GroganCVIU19}, \texttt{PCT\_OT} \cite{Alghamdi2019}, \texttt{OT\_NW} \cite{alghamdi2020patch} and our algorithms \texttt{NW\_c} and \texttt{NW\_cp}  (best viewed in colour and zoomed in). }
    \label{fig:qualitative_strips}
\end{sidewaysfigure*}

Figure  \ref{fig:qualitative_strips}  provides qualitative results. For  clarity, the results are presented in image mosaics, created by switching between the target image and the transformed source image column wise (Figure \ref{fig:qualitative_strips}, top row). If the colour transfer is accurate, the resulting mosaic should look like a single image (ignoring the small motion displacement between source and target images), otherwise column differences appear. 

\begin{figure*}
\centering
\begin{tabular}{c}
\includegraphics[width = .91\linewidth, height =.34\linewidth]{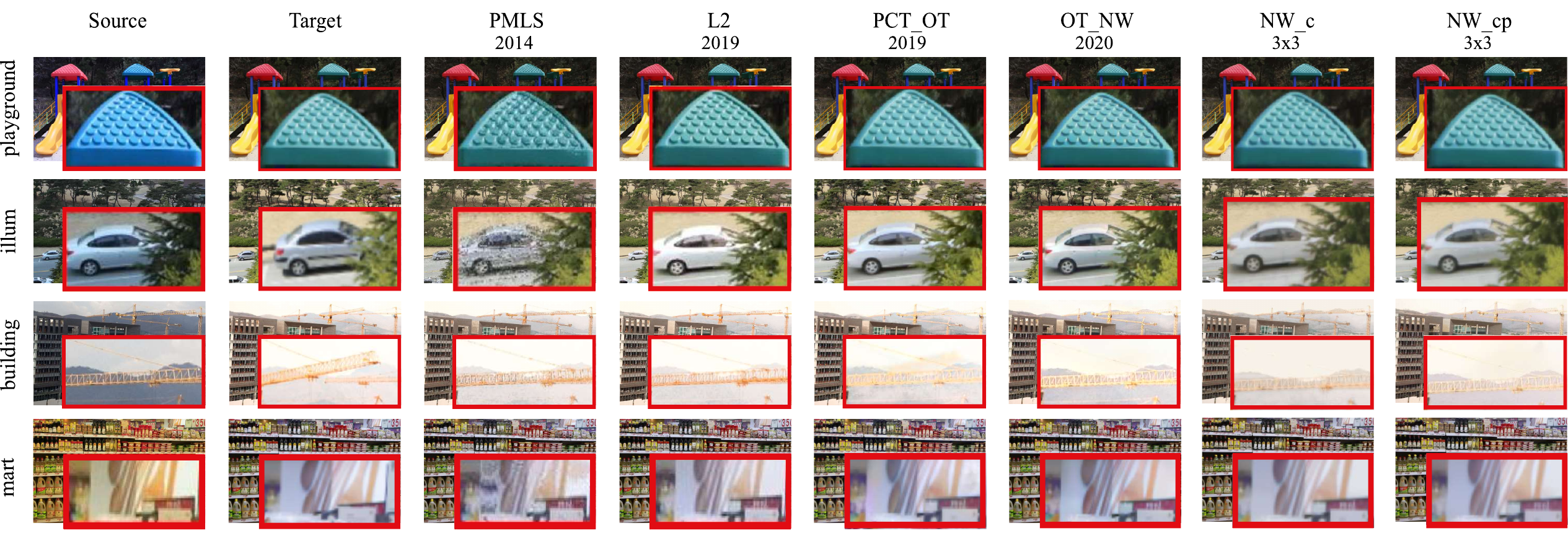}
\end{tabular}
\caption{A close up look at some of the results generated using the top performance methods  \texttt{PMLS} \cite{hwang2014color},  \texttt{L2} \cite{GroganCVIU19}, \texttt{PCT\_OT} \cite{Alghamdi2019} and \texttt{OT\_NW} \cite{alghamdi2020patch} and our algorithms \texttt{NW\_c} and \texttt{NW\_cp}  (best viewed in colour and zoomed in).}
\label{fig:artifacts}
  
\end{figure*}

While \texttt{PMLS} and \texttt{PCT\_OT} provide equivalent results to our method in terms of metrics measures, \texttt{PMLS} on the one hand introduces visual artifacts if the input images are not registered correctly (Figure \ref{fig:artifacts}), while our method is robust to registration errors. Note that although the accuracy of the PSNR, SSIM, CID and FSIMc metrics relies on the fact that the input images are registered correctly; if this is not the case, these metrics may not accurately capture all artifacts (Figures  \ref{fig:artifacts} and \ref{fig:qualitative_strips}). On the other hand, \texttt{PCT\_OT} can create shadow artifacts when there are large changes between target and source images (Figure \ref{fig:artifacts}, in example `building'), while our proposed methods with and without incorporating positions information can correctly transfer colours between images that contain significant spatial differences and alleviates the shadow artifacts, as can be seen in Figure \ref{fig:artifacts} with examples 'illum', 'mart' and 'building'. In addition, NW approach allows to create a smoother colour transfer result, and can also alleviate JPEG compression artifacts and noise (cf. Figure \ref{fig:artifacts} for comparison).


\section{Conclusion}

We have shown how to use the Nadaraya-Watson estimator to adapt  the IDT algorithm for accounting for input correspondences in registering  high dimensional probability density functions. Our approach is shown to be competitive to state of the art for colour transfer in images where spaces of dimension up to 45 have been used. Future work will look into combining solution $\phi^{OT}$ and $\phi^{NW}$ to tackle semi-supervised situations where correspondences are only partially available  \cite{Grogan:2017:UII:3150165.3150171,GroganCVIU19}.

\bibliographystyle{IEEEtran}
 \bibliography{IEEEabrv,ref}

\end{document}